\documentclass[pdflatex,sn-mathphys-num]{sn-jnl}

\usepackage{graphicx}%
\usepackage{multirow}%
\usepackage{amsmath,amssymb,amsfonts}%
\usepackage{amsthm}%
\usepackage{mathrsfs}%
\usepackage[title]{appendix}%
\usepackage{xcolor}%
\usepackage{textcomp}%
\usepackage{manyfoot}%
\usepackage{booktabs}%
\usepackage{algorithm}%
\usepackage{algorithmicx}%
\usepackage{algpseudocode}%
\usepackage{listings}%

\theoremstyle{thmstyleone}%
\theoremstyle{thmstyletwo}%

\theoremstyle{thmstylethree}%

\begin{document}

\title[Not Yet AlphaFold for the Mind: Evaluating Centaur as a Synthetic Participant]{Not Yet AlphaFold for the Mind: Evaluating Centaur as a Synthetic Participant}


\author[1]{\fnm{Sabrina} \sur{Namazova}}\email{snamazova@uni-osnabrueck.de}

\author[1]{\fnm{Alessandra} \sur{Brondetta}}\email{alessandra.brondetta@uni-osnabrueck.de}

\author[2]{\fnm{Younes} \sur{Strittmatter}}\email{ys5852@princeton.edu}

\author[3,4]{\fnm{Matthew} \sur{Nassar}}\email{matthew\_nassar@brown.edu}

\author*[1,5]{\fnm{Sebastian} \sur{Musslick}}\email{sebastian.musslick@uos.de}

\affil*[1]{\orgdiv{Institute of Cognitive Science}, \orgname{Osnabrück University}, \orgaddress{\city{Osnabrück}, \postcode{49090}, \country{Germany}}}

\affil*[2]{\orgdiv{Department of Psychology}, \orgname{Princeton University}, \orgaddress{\city{Princeton}, \postcode{08540}, \state{NJ}, \country{USA}}}

\affil[3]{\orgdiv{Robert J. and Nancy D. Carney Institute for Brain Science, Brown University}, \orgname{Brown University}, \orgaddress{\city{Providence}, \postcode{02906}, \state{RI}, \country{USA}}}

\affil[4]{\orgdiv{Department of Neuroscience}, \orgname{Brown University}, \orgaddress{\city{Providence}, \postcode{02906}, \state{RI}, \country{USA}}}

\affil[5]{\orgdiv{Department of Cognitive and Psychological Sciences}, \orgname{Brown University}, \orgaddress{\city{Providence}, \postcode{02912}, \state{RI}, \country{USA}}}

\maketitle


Simulators have revolutionized scientific practice across the natural sciences. By generating data that reliably approximate real-world phenomena, they enable scientists to accelerate hypothesis testing and optimize experimental designs \cite{jumper_highly_2021,krenn2016automated}. This is perhaps best illustrated by AlphaFold, a Nobel-prize winning simulator in chemistry that predicts protein structures from amino acid sequences, enabling rapid prototyping of molecular interactions, drug targets, and protein functions \cite{jumper_highly_2021}. In the behavioral sciences, a reliable participant simulator---a system capable of producing human-like behavior across cognitive tasks---would represent a similarly transformative advance \cite{Musslick2024}. Recently, Binz et al. introduced Centaur, a large language model (LLM) fine-tuned on human data from 160 experiments, proposing its use not only as a model of cognition but also as a participant simulator for ``in silico prototyping of experimental studies'' \cite{binz2025foundation}, e.g., to advance automated cognitive science \cite{Musslick2024,musslick2025automating}. Here, we review the core criteria for a participant simulator and assess how well Centaur meets them. Although Centaur demonstrates strong predictive accuracy, its generative behavior---a critical criterion for a participant simulator---systematically diverges from human data. This suggests that, while Centaur is a significant step toward predicting human behavior, it does not yet meet the standards of a reliable participant simulator or an accurate model of cognition.

A core criterion for any behavioral simulator is its ability to generate behavioral patterns observed in experiments. This includes reproducing established effects across diverse tasks and conditions and generalizing beyond known experiments. Such generalization is crucial if the simulator is to support hypothesis refinement, model comparison, or experimental design in unexplored domains. Importantly, a simulator need not provide a mechanistic account of natural phenomena. For example, AlphaFold accurately predicts protein structures from amino acid sequences without modeling the biophysical folding processes in vivo \cite{jumper_highly_2021}. Similarly, a participant simulator’s value lies in its capacity to produce realistic human-like behavior rather than explanatory insight. This contrasts with cognitive models, which aim to reveal the processes underlying behavior. However, reproducing human-like behavior remains critical for cognitive models: although they can offer theoretical insight, their scientific value is limited if they fail to account for behavioral phenomena \cite{gabaix2008seven}. Conversely, a predictive but opaque system---a black-box simulator---can still serve as a valuable synthetic participant, even without interpretability or mechanistic grounding.

\begin{figure}[h]
\centering
\includegraphics[width=1\textwidth]{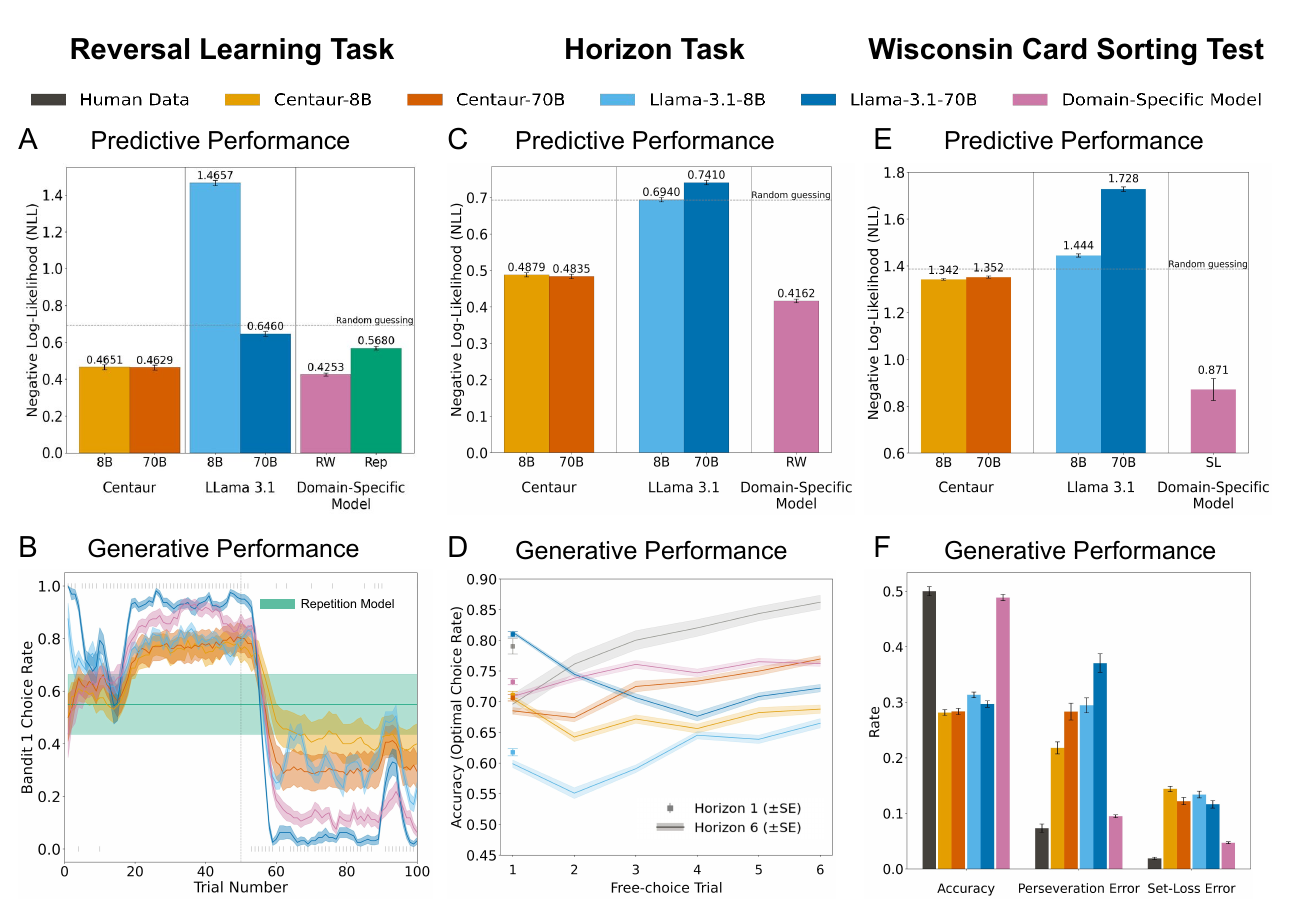}
\vspace{-0.5cm}
\caption{The small (7B) and large (80B) versions of Centaur were compared against respective Llama 3.1 models as well as domain-specific models. Domain-specific models included a Rescorla-Wagner (RW) model \cite{binz2025foundation}, a repetition (Rep) model \cite{palminteri2017importance}, and a sequential learning (SL) model \cite{bishara2010sequential} (see supplementary information). Models were evaluated for predictive and generative performance in a reversal learning task (A–B), a horizon-dependent bandit task (C–D), and the Wisconsin Card Sorting Test (E–F). Predictive performance was assessed via negative log-likelihood (NLL) fit to (A) synthetic data generated from the RW model in the reversal learning task \cite{eckstein2022reinforcement}, as well as to human data in the (C) horizon task \cite{wilson2014humans} and (E) Wisconsin Card Sorting Test \cite{steinke2020parallel}. (B) Generative performance in the reversal learning task shows the probability of choosing bandit 1 (initially rewarded at 80\% versus bandit 2 at 20\%) as a function of trial; a reward switch occurs on trial 50, reversing the reward probabilities. (D) In the horizon task, the proportion of optimal choices is plotted as a function of trials remaining, separated by phases with one free choice (horizon 1) versus six free choices (horizon 6). (F) Generative performance in the Wisconsin Card Sorting Test is shown as a function of error type (see main text). Error bars and shaded areas indicate the standard error of the mean across participants or simulated model instances.}\label{fig:results}
\end{figure}

A crucial distinction when evaluating simulators and computational models of behavior is the difference between predictive and generative performance \cite{steingroever2014absolute,palminteri2017importance}. A model may achieve high predictive accuracy by forecasting a participant’s next response based on their past responses, yet fail to generate plausible behavior when acting independently. In choice-based experimental tasks, predictive performance involves forecasting one trial at a time using previous choices and outcomes, whereas generative performance requires producing an entire sequence of choices de novo, without anchoring on the history of human choices. Consider a simple reversal learning task in which participants choose between two options (``bandits'') with different reward probabilities. Initially, one bandit is more rewarding, but at an unpredictable point, the reward probabilities reverse. Successful performance requires detecting this change and adapting behavior. A simple repetition model---one that merely repeats the prior choice with some probability---can achieve reasonable predictive accuracy for an adaptive participant, correctly forecasting most trials except the switch (Fig. \ref{fig:results}A) \cite{palminteri2017importance}. However, when run generatively, it is biased to repeat the same choice indefinitely, failing to adapt after the reversal and deviating qualitatively from human behavior (Fig. \ref{fig:results}B). A behavioral simulator, in contrast, must go beyond prediction, capturing key human features such as sensitivity to reward feedback and flexible adaptation when reward contingencies change.

While Binz et al. present evidence for Centaur’s predictive performance \cite{binz2025foundation}, its ability to generate behavior---a hallmark of both cognitive models and behavioral simulators---remains largely untested. Here, we evaluate both predictive and generative performance across three tasks. The first is a reversal learning task, part of the bandit task class Centaur was trained on \cite{eckstein2022reinforcement}, where participants must adapt choices after an unannounced switch in reward contingencies. Humans and animals reverse their choices several trials after such a switch \cite{eckstein2022reinforcement} \cite{izquierdo2017neural}. The second task is a horizon-dependent bandit task, also within Centaur’s training set, where participants balance exploration and exploitation under varying time horizons \cite{wilson2014humans}. In this tasks, humans show a horizon effect, foregoing the highest expected value action more frequently in favor of exploration in situations that involve a longer trial horizon over which to exploit knowledge gained. The third task is the Wisconsin Card Sorting Test \cite{steinke2020parallel}, which Centaur was not fine-tuned on. Here, participants must infer and apply an unstated card selection rule (e.g., according to color, shape, or number of items displayed on the card) based on feedback, then flexibly switch when the rule changes unexpectedly. Human participants generally perform well on the task but exhibit errors because they fail to adapt to a new rule (perseveration error) or to maintain the correct rule (set-loss error).  

Fig. \ref{fig:results} summarizes Centaur’s predictive and generative performance across the three tasks, benchmarked against respective base LLMs (Llama-3.1-8B, Llama-3.1-70B) and domain-specific models (see Supplementary Information). In the reward reversal learning task, we observe a stark dissociation between predictive and generative performance. Centaur surpasses Llama 3.1-70B in predictive performance (Fig. \ref{fig:results}A), but exhibits weaker reversal dynamics when run generatively (Fig. \ref{fig:results}B). Notably, some seeds result in Centaur failing to adapt its choice behavior altogether following the reward reversal (Fig. \ref{fig:bandit_prop_across_seeds}). In contrast, Llama 3.1-70B, despite lower predictive accuracy, produces generative behavior that more closely resembles the characteristic reversal seen in the data-generating model. A similar pattern emerges in the horizon-dependent bandit task that Centaur was fine-tuned on. While Centaur’s predictive performance is comparable to that of the domain-specific model (Fig. \ref{fig:results}C), its generative behavior deviates substantially from human data, failing to capture the effects of the horizon manipulation or a comparable increase in proportion of optimal choices with time characteristic of a switch from exploration to exploitation (Fig. \ref{fig:results}D). Finally, in the Wisconsin Card Sorting Test---a task outside Centaur’s fine-tuning set---the domain-specific model outperforms Centaur on both predictive (Fig. \ref{fig:results}E) and generative measures (Fig. \ref{fig:results}F).  Centaur does not achieve human-like accuracy on the task, exhibiting substantially more perseveration and set-loss errors. 

Our findings suggest that while Centaur achieves high predictive performance on tasks it was trained on, it still struggles to reproduce human-like behavior in those tasks, including the qualitative hallmarks of behavior that the tasks themselves were designed to measure (i.e., reversals, horizon effects). Moreover, on the task outside its fine-tuning set, it performed worse than the domain-specific model. These limitations constrain its utility both as a synthetic participant, e.g. to prototype novel experimental designs \cite{Musslick2024}, and as a model of cognition. To fulfill these roles, it must not only predict trial-by-trial choices based on human histories but also reliably generate behavior de novo. Insights from simulators like AlphaFold point to promising paths forward, such as integrating mechanistic constraints or developing standardized benchmarks to assess generative performance. Centaur represents a significant step toward aligning LLMs with human behavior, but predictive performance alone is insufficient for it to be useful as a participant simulator. Without the ability to simulate behavior faithfully, Centaur---and similar models---cannot yet serve as a ``behavioral AlphaFold.''


\section*{Data and Code Availability}
Data and code for all reported analyses all figures are available at \href{https://github.com/snamazova/centaur_evaluation}{https://github.com/snamazova/centaur\_evaluation}.

\section*{Competing Interests}
The authors declare no competing interests.

\section*{Author Contributions}
\textbf{Project lead:} Sebastian Musslick\\
\textbf{Data curation:} Sabrina Namazova, Alessandra Brondetta\\
\textbf{Model evaluation and analysis:} Sabrina Namazova, Alessandra Brondetta, Younes Strittmatter\\
\textbf{First draft}: Sebastian Musslick\\
\textbf{Review and editing:} All authors\\
We thank Marcel Binz for valuable feedback on using and evaluating Centaur. We also thank Moritz Hartstang and Cornelius Wolff for helpful discussions.

\backmatter



\section*{Supplementary Information}

\subsection*{Models and Data}
The small (7B) and large (80B) versions of Centaur were compared against their respective Llama-3.1-Instruct models as well as domain-specific cognitive models. To ensure consistency and fair comparison, all models were evaluated on the same test sets for each task. Human behavioral datasets, taken from prior studies, or synthetic data were used to fit the domain-specific models and evaluate predictive and generative model performance. 

Below, we outline general information about the fitting procedure as well as information about task design, data, domain-specific models specific to each task.

\subsection*{Predictive Performance Evaluation}
In alignment with Binz et al. \cite{binz2025foundation}, we examined the extend to which each model predicts the choices of participants in the evaluation datasets, conditioned on their previous history of choices in the experiment. We note that this procedure places traditional models at a disadvantage, as they are typically fitted to individual participants and evaluated on hold-out trials from the same participant. Nevertheless, for consistency, we followed the evaluation approach described by Binz et al.

\textbf{Prompt Construction.} For LLMs, the prompt at the first trial includes only the task instructions and the first stimulus. After every trial, we append the participant's choice and the feedback it received. Thus, by trial $t$, the prompt reflects the exact sequence of choices and feedback observed by the participant up to trial $t – 1$. The prompt is reset at the beginning of each new participant to prevent any information leakage across participants. For the reversal learning task and the horizon task, we used the prompts from the corresponding experiments included in the Centaur training set \cite{binz2025foundation}. The prompts for the Wisconsin Card Sorting Test are described in the relevant task section below.

\textbf{Evaluation.} Following Binz et. al \cite{binz2025foundation}, models' goodness-of-fit is reported as the average negative log-likelihood (NLL) of participant responses across all evaluation trials. Specifically, for each trial, we extract the model’s predicted logits at the final token position, apply a softmax over the full vocabulary (or over the constrained answer set in the case of domain-specific models), and compute the log-probability assigned to the participant's actual choice. The loss per trial is then recorded as $\mathrm{NLL} = -\log p(\text{participant choice})$ and averaged across all trials in the evaluation set:
$\text{mean NLL} = -\frac{1}{T} \sum_{t=1}^{T} \log p_t(\text{participant choice}_t)$.
Lower NLL values (i.e., higher assigned probabilities) indicate better predictive model performance.

\subsection*{Generative Performance Evaluation}
To probe Centaur's utility as a participant simulator, we evaluated whether the models were able to generate human-like behavior and task-relevant statistics in an open-loop (generative) setting. 

\textbf{Prompt Construction.} The prompt configuration follows that of the predictive performance evaluation: it begins with the task instructions and the first stimulus. However, from the first trial onward, we add the model’s own selected choice and the corresponding feedback to the prompt. As a result, the model's internal context evolves exclusively on the basis of its own behavior. To ensure reproducibility and independence across runs, the prompt and all random number generators are re-initialized and reseeded at the start of each simulated participant.

\textbf{Action Policy.} The models generate a response on each trial by sampling their softmax output distribution (temperature = 1), allowing behavioral variability between simulated participants.

\textbf{Evaluation Metrics.} For each task, we computed task-specific summary statistics and behavioral markers from the model-generated data and compared them against the observed behavior, as described for each task below.

\subsection*{Reversal Learning Task}

\textbf{Task description.} 
On each trial of this two-armed bandit task, participants choose between two options (``bandits'') to select the one that yields a reward. The reward is provided probabilistically, with one bandit yielding a higher chance of reward than the other. This task involves a deterministic reversal in reward contingencies for the two bandits. For the first 50 trials, bandit~1 delivers a reward with 80\% probability, and bandit~2 with 20\%. In the second half, the reward probabilities are reversed: bandit~2 is rewarded 80\% of the time, and bandit~1 only 20\%. 

This task design was inspired by a reward reversal learning task with multiple reversals included in the training set of Centaur \cite{eckstein2022reinforcement}. However, here, we simulated a single reward reversal to better isolate the behavioral adaptation to the reversal, using the same 2-armed bandit prompts as described in \cite{binz2025foundation}.

\textbf{Domain-Specific Model.}
As the primary domain-specific model, we used the Rescorla--Wagner (RW) model reported by Binz et al. \cite{binz2025foundation}. Although the original formulation included five free parameters, we simplified it to three free parameters to avoid overparameterization, which can harm predictive performance.
\begin{itemize}
    \item $\alpha$: learning rate
    \item $\beta$: inverse temperature
    \item $d$: initial value estimate for each bandit
\end{itemize}

Action probabilities were computed using the softmax function:
\[
\ p_i = \frac{\exp(\beta \cdot V_i)}{\sum_{j=1}^{N} \exp(\beta \cdot V_j)}
\]

Value updates followed the standard RW rule:
\[
\delta_t = r_t - V_{c_t}
\]
\[
V_{c_t} \leftarrow V_{c_t} + \alpha \delta_t
\]

Following the rationale introduced by Palminteri et al. \cite{palminteri2017importance}, we considered a repetition model as a second domain-specific baseline for this task. The repetition model was designed to capture perseverative behavior by repeating the previous choice with a certain probability $p$, and selecting the alternative choice with probaility $1-p$. The probability of repeating the previous choice was treated as a free parameter and was estimated by minimizing the NLL. As outlined in the main text, the repetition model illustrates the difference between predictive and generative performance: While the model can achieve reasonable predictive performance on the reversal learning task (Fig. \ref{fig:results}A), its generative performance does not match human behavior (Fig. \ref{fig:results}B). 

\textbf{Data.} 
For this task, we generated synthetic data from the RW model described above using the following parameters: $\alpha = 0.5$, $\beta = 2.5$, and $d = 0.5$. Synthetic data was generated by running the model on the reversal learning task. To account for variability, data was generated using 32 different random seeds.

\textbf{Predictive Performance Evaluation.} 
To assess predictive performance, the RW model was evaluated on data it had generated itself by minimizing NLL. The resulting average NLL was used as a baseline for comparison. The repetition model was also evaluated on the RW-generated data: its probability of repeating the previous action was fitted to this dataset, and the same evaluation procedure with averaging NLL as for the RW model was applied. Centaur-7B and Centaur-80B were then tested on the same synthetic data (generated using the RW model), using the evaluation procedures outlined above. Llama-7B and Llama-80B Instruct models were also evaluated to assess the effect of instruction fine-tuning on predictive performance. Using the same evaluation setup described above, we computed the average NLL across trials, where only the token corresponding to the human choice was unmasked during loss computation.

\textbf{Generative Performance Evaluation.}
To assess generative performance, all models were run on the same stimulus timeline across 32 random seeds. For the repetition model, a random bandit was selected at the start of each run, and the model repeatedly chose this same arm throughout the timeline with probability $p$. For the LLMs, simulations were conducted in an open-loop manner as described above. We recorded the proportion of trials in which bandit 1 was chosen. 

We evaluated models based on a reversal effect, reflecting a decrease in the proportion of times bandit 1 (the no longer rewarded option) was chosen. For Centaur, we observed a weak reversal effect with large variability (Fig. \ref{fig:results}B). To further examine Centaur-70B’s generative behavior, we examined individual random seeds (Fig \ref{fig:bandit_prop_across_seeds}). For some seeds, Centaur exhibits a proportion of 1 for choosing bandit 1, suggesting that it consistently repeats the same choice after the initial trials. This indicates that certain seeds fail to show the reversal learning effect entirely. Additionally, we observe that some seeds display a fluctuating choice proportion for bandit 1 (between 0.4 and 0.6), suggesting that the model may be switching between options in a way that reflects random behavior rather than reward-based decision-making. 

\renewcommand{\thefigure}{S\arabic{figure}}
\setcounter{figure}{0}

\begin{figure}[H]
\centering
\includegraphics[width=0.75\textwidth]
{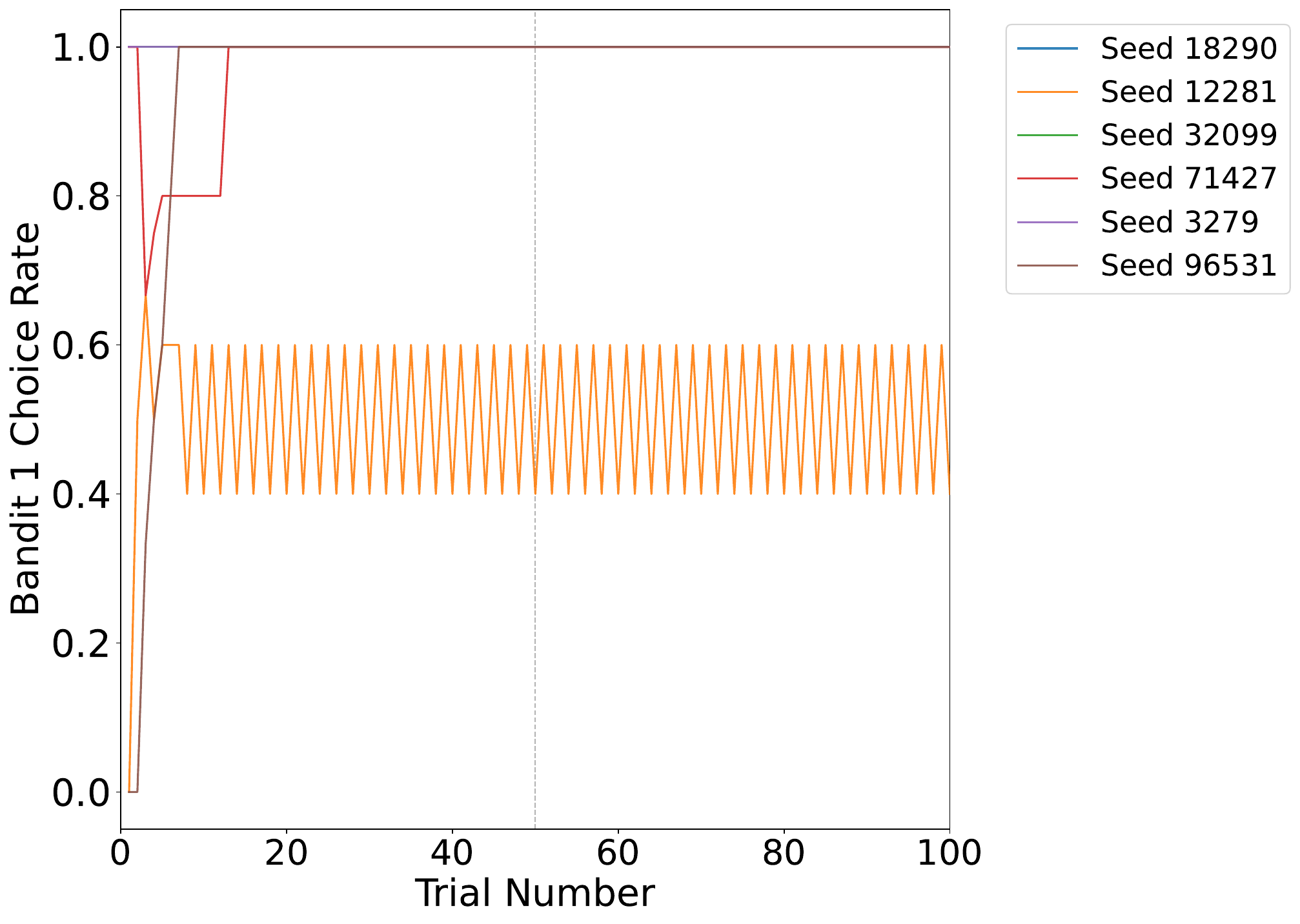}
\caption{Probability of choosing bandit 1 across trials in the reversal learning task, illustrated for selected initializations of Centaur. Each line corresponds to a specific random seed. The vertical line at trial 50 indicates the reward reversal, at which bandit 1 is no longer more rewarded than bandit 2.}
\label{fig:bandit_prop_across_seeds}
\end{figure}

\subsection*{Horizon Task}

\textbf{Task Description.}
The horizon task is a variant of a two-armed bandit task that incorporates both instructed and free-choice phases \cite{wilson2014humans}. Each phase begins with four forced-choice trials, during which participants are instructed which bandit to select. These are followed by either one (horizon 1) or six (horizon 6) free-choice trials, where participants can choose between the two bandits freely. The manipulation of horizon length (horizon 1 vs. horizon 6) is intended to investigate how participants balance exploration---gathering information about uncertain options---and exploitation—choosing the option believed to yield the highest reward. Here, we implemented the task as described in Experiment 1 of \cite{wilson2014humans}. For LLM evaluations we used the prompts described in \cite{binz2025foundation}.

\textbf{Domain-Specific Model.}
As the domain-specific model, we used the Rescorla--Wagner (RW) model described in the reversal learning task and in Binz et al. \cite{binz2025foundation}.

\textbf{Data.}
We used the human behavioral data from the original study \cite{wilson2014humans}.

\textbf{Predictive Performance Evaluation.}
To assess predictive performance, the original dataset (31 participants) was split in an 80/20 ratio: 25 participants were used for training and 6 participants for testing. The RW model was fit to the training data. Model parameters were estimated by minimizing NLL on free-choice trials; although model parameters were also updated using forced-choice trials, NLL was not computed for these. After fitting, data were simulated using the fitted parameters, and NLL was computed between the simulated and actual free-choice trials in the test set. The averaged NLL from this process was reported as the model's predictive performance.

Centaur-7B and Centaur-80B were evaluated using the evaluation procedures described above. However, the negative log-likelihood (NLL) was computed using only free-choice trials; therefore, although the model was informed during instructed choices, predictions and NLL calculations were performed exclusively on free-choice trials.

The Llama-7B and Llama-80B Instruct models were also evaluated to assess the effect of instruction fine-tuning on predictive performance. Predictive performance was evaluated as in the reversal learning task by unmasking the human choice and computing the loss. As with Centaur, only free-choice trials were considered.

\textbf{Generative Performance Evaluation.}
To assess generative performance, timelines for all 31 participants were extracted. Each random seed was mapped to a specific participant’s timeline and used to simulate model behavior. Here, we sought to replicate the effects reported in Fig.~1c of the original study \cite{wilson2014humans}, assessing the optimal choice rate as a function of the horizon condition and trials left to perform. The optimal choice rate was calculated as the proportion of choices that were optimal, where the bandit with the higher generative mean was considered optimal. Optimal choice rates were plotted across free-choice trials within the horizon-6 condition.


\subsection*{Wisconsin Card Sorting Test}

\textbf{Task Description.} 
The Wisconsis Card Sorting Test (WCST) is one of the most widely used neuropsychological assessments of cognitive flexibility and executive functioning \cite{grant1948behavioral,nyhus2009wisconsin}. 
In this task, participants must sort stimulus cards to one of four key cards based on a hidden category (either color, form, or number) that changes periodically throughout the session. On each trial, participants receive feedback indicating whether the currently applied category is correct or incorrect. The goal is to infer the correct matching rule from the feedback and adapt the sorting behavior accordingly. 

Key behavioral metrics include the proportion of trials sorted according to the correct rule (accuracy), repeating the incorrect category following negative feedback (perseveration error), and switching the correct category following positive feedback (set-loss error). These metrics reflect success or failure in trial-by-trial, feedback-driven category learning. In the generative evaluations (Fig. \ref{fig:results}F), we use these metrics to compare the models' generative behavior against human behavior.

\textbf{Domain-Specific Model.}
For the WCST, we implemented a four-parameter sequential learning (SL) model from the family of models proposed by Bishara et al. \cite{bishara2010sequential}. The SL model assumes participants maintain an internal attention vector $\mathbf{a}(t)$ over the three sorting dimensions (color, form, number), representing the subjective relevance of each rule. This attention vector is updated on a trial-by-trial basis in response to a feedback signal $\mathbf{s}(t)$, and it affects the probability of selecting each key card $P_k(t)$, guiding future choices. The four parameters of the model are:
\begin{itemize}
    \item $r$: the learning rate for positive feedback (“REPEAT”), controlling how quickly attention weights change in response to rewarding feedback; $r \in [0, 1]$.
    \item $p$: the learning rate for negative feedback (“SWITCH”), controlling how quickly attention weights change in response to punishing feedback; $p \in [0, 1]$.
    \item $d$: the decision consistency, implemented as a softmax inverse-temperature; higher values lead to more deterministic choices; $d \in [0.01, 5]$.
    \item $f$: the attentional focus parameter, modulating how sharply attention affects learning; although $f$ plays a limited role here due to unambiguous feedback; $f \in [0.01, 5]$.
\end{itemize}

The 3-dimensional attention vector $\mathbf{a}(t)$ on trial $t$:

\begin{equation}
\mathbf{a}(t) = 
\begin{pmatrix}
a_{\text{color}}(t) \\
a_{\text{form}}(t) \\
a_{\text{number}}(t)
\end{pmatrix}
\end{equation}

In trials with positive feedback, the feedback signal for category $i$ is computed as:

\begin{equation}
s_i(t) \Big|_{\text{positive}} = 
\frac{m_{k,i}(t)\, [a_i(t)]^f}
{\sum_{h=1}^{3} m_{k,h}(t)\, [a_h(t)]^f}
\end{equation}

In trials with negative feedback, the signal is:

\begin{equation}
s_{i}(t) \Big|_{\text{negative}} = 
\frac{(1 - m_{k,i}(t)) \, [a_{i}(t)]^f}
{\sum_{h=1}^{3} \left[(1 - m_{k,h}(t)) \, [a_{h}(t)]^f\right]}
\end{equation}

Here, $\mathbf{m}_i(t)$ is a 3-dimensional match vector that indicates whether each category $i$ matches between the stimulus and the selected key card $k$ on trial $t$: $m_{k,i}(t) = 1$ if matched, and $0$ otherwise.

The attention vector is then updated by integrating the feedback with a learning rate $r$, following positive feedback, or $p$, following negative feedback:

\begin{equation}
\mathbf{a}(t+1)\Big|_{\text{positive}} = (1 - r)\, \mathbf{a}(t) + r\, \mathbf{s}(t)
\end{equation}
\begin{equation}
\mathbf{a}(t+1)\Big|_{\text{negative}} = (1 - p)\, \mathbf{a}(t) + p\, \mathbf{s}(t)
\end{equation}

Finally, the probability of choosing key $k$ on trial $t$ is given by:

\begin{equation}
P_k(t) = \frac{ \mathbf{m}_k^\top(t)\, \left[ \mathbf{a}(t) \right]^d}
{\sum_{j=1}^{4}\mathbf{m}_j^\top(t)\,  \left[  \mathbf{a}(t) \right]^d}
\end{equation}

\textbf{Data.} 
Human data was taken from an existing dataset comprising a large sample of undergraduates (375 participants) who completed a computerized variant of the WCST (cWCST) \cite{steinke2020parallel}. In this version, the four key cards were fixed across all trials, and each of the 24 stimulus cards shared at most one feature with any key card, ensuring that the correct category was unambiguous. Participants responded via keyboard and received visual feedback (``REPEAT'' or ``SWITCH'') in place of standard ``correct'' or ``incorrect'' cues. Each participant was required to complete 41 correct rule switches, with a maximum of 250 trials.

\textbf{Example Prompt.} 
In the LLM simulations, the cWCST was implemented as a sequential dialogue in natural language. The initial prompt includes the task instructions and definitions of the four key cards. In each trial, the model is shown a stimulus card and asked to select a matching key (A–D). Below is an example prompt (from the fourth trial) used in the Centaur simulations: \vspace{\baselineskip}

\begin{quote}
You will see a stimulus card and must choose which of four key cards it matches. Cards can be matched by one of three categories: color, form, or number. The matching category changes from time to time. 
After each choice, you will receive a feedback stimulus:

- ``REPEAT'' means you used the correct category and should keep using it.

- ``SWITCH'' means you used the wrong category and should try a different one.

The four key cards are always:

- A: one red triangle

- B: two green stars

- C: three yellow crosses

- D: four blue balls

Each stimulus card shares at most one property (color, form, or number) with any one key card. Your task is to use the feedback to figure out the correct temporary category to apply and respond accordingly pressing key 'A' or 'B' or 'C' or 'D'.

You see the following stimulus card: one blue cross. You press key $<<$ A $>>$ (one red triangle). You get the following feedback stimulus: SWITCH.

You see the following stimulus card: four yellow triangle. You press key $<<$ C $>>$ (three yellow cross). You get the following feedback stimulus: SWITCH.

You see the following stimulus card: three red star. You press key $<<$ B $>>$ (two green star). You get the following feedback stimulus: REPEAT.

You see the following stimulus card: three red ball. You press key $<<$
\end{quote}

\vspace{\baselineskip}

\textbf{Predictive Performance Evaluation.} 
To asses predictive performance, the human dataset was split into a 80/20 ratio. The parameters of the domain-specific sequential learning model were fit on 80\% of participants of the dataset by minimizing the total trial-wise negative log-likelihood (NLL) across their choices. The fitted model was then evaluated on the remaining 20\% of participants. We fitted the domain-specific model globally across participants, meaning all participants share the same parameter values: $r=0.967, p=0.656, d=0.41, f=0.05$. For predictive performance (Fig. \ref{fig:results}E), all models, including Centaur, Llama, and the domain-specific model, were evaluated on the same 20\% held-out participants, following the procedure described above.

\textbf{Generative Performance Evaluation.} 
To asses generative performance, stimulus timelines for the same 20\% held-out human participants were used to simulate model behavior in an open-loop setting, in which models received feedback based on their own choices rather than human responses. The generative procedure for all models followed the generative performance evaluation protocol described above. To evaluate how closely each model captured human-like behavior, we computed accuracy, perseveration error rate, and set-loss error rate, averaged across participants. These metrics were then also compared against the corresponding distributions observed in the human data (cf. Fig. \ref{fig:results}F).




\newpage

\bibliography{sn-bibliography}

\end{document}